\documentclass{article}

    \usepackage[preprint]{neurips_2023}

\usepackage[utf8]{inputenc} % allow utf-8 input
\usepackage[T1]{fontenc}    % use 8-bit T1 fonts
\usepackage{hyperref}       % hyperlinks
\usepackage{url}            % simple URL typesetting
\usepackage{booktabs}       % professional-quality tables
\usepackage{amsfonts}       % blackboard math symbols
\usepackage{nicefrac}       % compact symbols for 1/2, etc.
\usepackage{microtype}      % microtypography
\usepackage{xcolor}         % colors
\usepackage{algorithm}
\usepackage{algpseudocode}

\definecolor{myred}{RGB}{200, 0, 0}
\usepackage{pifont}

\usepackage{amsmath}
\usepackage{amssymb}
\usepackage{mathtools}
\usepackage{amsthm}

\usepackage{graphicx}
\usepackage{caption}
\usepackage{subcaption}

\renewcommand{\[}{\begin{eqnarray}}
\renewcommand{\]}{\end{eqnarray}}

\usepackage{wrapfig}

\hypersetup{
    colorlinks,
    linkcolor={red!50!black},
    citecolor={blue!50!black},
    urlcolor={blue!80!black}
}

\usepackage[capitalize,noabbrev]{cleveref}

\theoremstyle{plain}

\theoremstyle{definition}

\theoremstyle{remark}

\newcommand\norm[1]{\left\lVert#1\right\rVert}

\DeclareMathOperator{\ent}{Ent}

\DeclareMathOperator{\E}{\mathbb{E}}

\title{Layer Collaboration in the Forward-Forward Algorithm}

\author{%
  Guy Lorberbom$^1$\thanks{Equal contribution}
  \And
  Itai Gat$^{2\text{ }*}$
  \And
  Yossi Adi$^3$
  \And
  Alex Schwing$^4$
  \And
  Tamir Hazan$^1$
  \AND
    \\
    \textsuperscript{\rm 1} Technion\\
    \textsuperscript{\rm 2} FAIR Team, Meta AI Research\\
    \textsuperscript{\rm 3} The Hebrew University of Jerusalem\\ \textsuperscript{\rm 4}University of Illinois at Urbana-Champaign\\
}

\begin{document}

\maketitle

\begin{abstract}
    Backpropagation, which uses the chain rule, is the de-facto standard algorithm for optimizing neural networks nowadays. Recently,~\citet{ff} proposed the forward-forward algorithm, a promising alternative that optimizes neural nets layer-by-layer, without propagating gradients throughout the network. Although such an approach has several advantages over back-propagation and shows promising results, the fact that each layer is being trained independently limits the optimization process. Specifically, it prevents the network's layers from collaborating to learn complex and rich features. In this work, we study layer collaboration in the forward-forward algorithm. We show that the current version of the forward-forward algorithm is suboptimal when considering information flow in the network, resulting in a lack of collaboration between layers of the network. We propose an improved version that supports layer collaboration to better utilize the network structure, while not requiring any additional assumptions or computations. We empirically demonstrate the efficacy of the proposed version when considering both information flow and objective metrics. Additionally, we provide a theoretical motivation for the proposed method, inspired by functional entropy theory.
\end{abstract}

\section{Introduction}

Deep neural networks (DNNs) have been the backbone of many state-of-the-art machine learning systems in recent years. Their ability to learn complex representations from large amounts of data has led to breakthroughs in many fields such as image and speech recognition, natural language processing, and game playing~\citep{lavin2016fast, bojarski2016end, luss2019generating, NEURIPS2021_b51a15f3, 9747870, rad, ophir2020deep, zhou2021review, Hurtley645, fedorov2020tinylstms}. One of the key reasons for the success of deep neural networks is the backpropagation algorithm, which is currently the most widely used training paradigm.

Backpropagation~\citep{rumelhart1986learning} utilizes the chain-rule to compute gradients for minimization of a loss function, which measures the discrepancy between the predicted output and the desired result. The algorithm starts with a forward pass through the network to compute the predicted output and then calculates the error between the prediction and the desired output. The error is then propagated back through the network using the chain rule of calculus, computing the gradient of the loss function with respect to the network's parameters, i.e., the weights and biases. These gradients are then used to update the parameters via gradient descent variants, reducing the loss function and subsequently the error. This process is repeated until the network reaches a satisfactory error on the training data. The chain rule presented challenges for a variety of model architectures. In very deep networks, gradients of the loss function with respect to the weights can become small, resulting in vanishing gradients. Moreover, the gradient can also become too large. Both vanishing and exploding gradients lead to poor performance.

Recently,~\citet{ff} introduced the forward-forward algorithm, which offers an alternative to the backpropagation training process by replacing the forward and backward passes with two forward passes. The data that is forwarded in the network is the sample itself (e.g., image, text) concatenated with a one-hot encoded vector that represents a  label. The first forward pass is done with positive data (i.e., data concatenated with the true label) and the second forward pass is done with negative data (i.e., data concatenated with a false label). The network then classifies whether a sample is concatenated with its true label. An important aspect of the forward-forward algorithm: the aforementioned optimization process is performed separately for each layer.

\begin{figure*}[t!]
     \centering
     \includegraphics[width=0.9\textwidth]{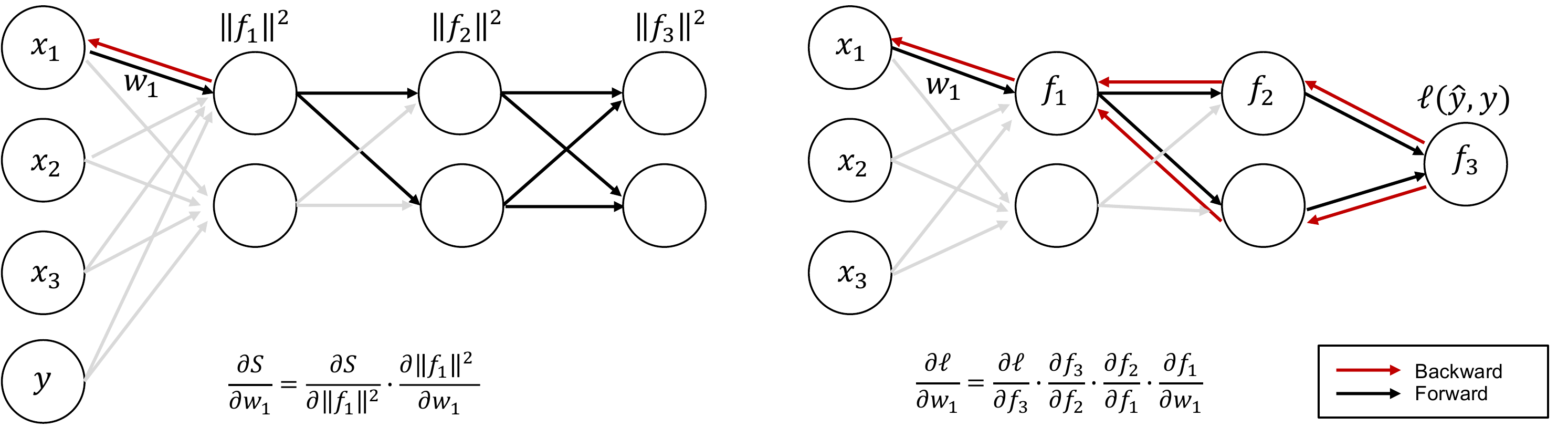}
     \caption{Information flow for deep net training with the forward-forward (left) and backpropagation  (right)   algorithm when focusing on one branch. The forward pass in backpropagation transfers information from layers to their successors, and the backward pass transfers information from later layers to their predecessors. The forward-forward algorithm feeds information to later layers from its predecessors in the forward pass. However, they are unaware of their successors. Hence, the ability of the forward-forward algorithm to construct hierarchical representations is limited. $S$ denotes the sigmoid function.}
     \label{fig:motivation}
\end{figure*}

In this work, we examine the dynamics of individual layers during the forward-forward algorithm based training process. Our findings indicate that, during training, the layers exhibit a lack of effective intercommunication, which results in a suboptimal level of collaboration. Based on this observation, we propose a simple modification to the forward-forward algorithm to enhance inter-layer communication during training. Our empirical results demonstrate that incorporating this modification leads to a significant improvement in performance.

To further investigate this phenomenon, we analyze forward-forward-based models through the lens of functional entropy. We demonstrate that the forward-forward method implicitly maximizes the entropy within the network. Throughout this work, we analyze the dynamics of the entropy when training a model using the forward-forward algorithm.

\paragraph{Our contributions:} We begin by analyzing the layer collaboration of forward-forward-based models. We find that those models are limited in their ability to transfer information during training and consequently exhibit a poor collaboration between layers. Then, we propose a modification to the forward-forward algorithm that enables better information flow through the network. We show the effectiveness of this modification in terms of performance and layer collaboration. Our work also presents a theoretical connection between the forward-forward algorithm and functional entropy. Using it, we are able to investigate the dynamics of the layer through the lens of entropies.

The remainder of the work is organized as follows: Section~\ref{sec:background} provides details regarding the forward-forward algorithm and how it differs from the popular backpropagation method.  Section~\ref{sec:layer_colb} analyzes the collaborative dynamics of the layers of forward-forward-based models and  proposes a modification. Section~\ref{sec:entropy} connects forward-forward optimization and the functional entropy of the layers. In Section~\ref{sec:experiments} we extensively study the benefits of  our proposed modification using performance metrics and our proposed theoretical framework. Lastly, Section~\ref{sec:related} covers related works in the areas of layer collaboration and neural-network learning methods.

\section{Background}\label{sec:background}

Discriminative learning seeks to construct a mapping between data-instance $x\in\mathcal{X}$ and corresponding label $y\in\mathcal{Y}$, given training data $\{(x_1, y_1),...,(x_m, y_m)\}$. Classical discriminative deep fully-connected neural networks have a recursive structure. Each layer processes its predecessor's output. Thus, a \emph{k}-layer fully connected network $f$ can be viewed as a chain of its sub-functions, 
\[
    f_{1:k}(x) \triangleq (f_k \circ ... \circ f_1)(x),
\]
where each layer $i$ is denoted by $f_i$. In the following, we cover backpropagation and forward-forward algorithms. In this work, we also point out some of their differences, as well as pros and cons.

\subsection{Backpropagation}

Backpropagation-based learning optimizes the parameters $w$ of a network $f$  given a set of training data points and a pre-defined loss function $\ell(f, x, y)$. During tuning, one or multiple data points are forwarded through the network to arrive at a prediction. The loss considers the output of the network and uses it to quantify the performance of the network. Formally, the optimization seeks to address the program:
\[
    \min_{w} \ell(f_{1:k}, x, y).
\]
This program is often addressed by gradient-based methods, such as stochastic gradient descent (SGD) and its variants. For example, the optimization step for the \emph{i}-th layer is:
\[
    \Delta w_{i} = - \eta \frac{\partial \ell\left(f_{1:k}, x, y\right)}{\partial w_{i}}= - \eta \frac{\partial \ell\left(f_{1:k}, x, y\right)}{\partial f_{1:k}(x)}\cdot\ldots\cdot\frac{\partial f_{1:i+1}(x)}{\partial f_{1:i}(x)}\frac{\partial f_{1:i}(x)}{\partial w_{i}}.\label{eq:backprop}
\]
Consequently, during backpropagation-based learning, information first flows from the input to the output in the forward pass. Conversely, in the backward pass, information flows from the output to the input (see Figure~\ref{fig:motivation}).

\subsection{The Forward-Forward algorithm}
The forward-forward algorithm provides an alternative method to learn deep neural network parameters. Different from backpropagation, the samples are represented as the raw input data point $x$ linked with a one-hot encoding vector that points to a label $y$. Then, it splits the training data into `positive' and `negative' sets. The one-hot vector in the positive samples is pointing at the true label. For the negative samples, the label vector is pointing at a randomly chosen wrong label. Intuitively, it seeks to learn whether a linked sample is positive or negative.

The main idea behind the forward-forward algorithm: make each layer `excited' about positive samples and, at the same time,  make each layer less excited about negative samples. Formally, given a data point linked with a label, we seek to classify whether this set is positive or negative based on the goodness of the network. Let us define the goodness of the \emph{i}-th layer as its squared sum of ReLU activities. The probability of a sample $(x, y)$ to be positive given by the \textit{i}-th layer  is computed via
\[
    p_{i}(\text{positive}) \triangleq \left(1 + e^{-(\norm{f_{1:i}(x,y)}^2 - \theta)} \right)^{-1},
\]
where $\theta$ is a hyperparameter that controls the scale of the gradient. Normalization is applied between layers such that later layers aren't able to rely on their predecessor's decision.

Finally, to perform learning, we optimize each layer sequentially to maximize goodness on positive samples and minimize negative ones. Namely, we train each layer once until convergence and then optimize its successor. Using SGD for layer optimization, we obtain a positive optimization step for the \emph{i}-th layer via
\[
    \Delta w_i = \frac{\partial \log\left(\left(1 + e^{-(\norm{f_{1:i}(x,y)}^2 - \theta)} \right)^{-1}\right)}{\partial w_i}.\label{eq:ff_step}
\]
Similarly, the negative step is performed with the opposite sign. Throughout the paper, we also refer to this loss as `Layer loss.' Different from backpropagation, this allows an individual training process for each layer, and, as a result, one can learn layers with non-differentiable functions in between the optimized layers, such as solvers~\citep{indelman2021learning}, argmax operation~\citep{lorberbom2019direct}, etc. Notice, similarly to energy-based models~\citep{lecun2005loss}, the forward-forward algorithm assigns a score to every $(x, y)$ pair. However, instead of training the network using backpropagation, the network is trained layer-by-layer.

Different from the training step, the inference process requires assessing the goodness of the network for each possible label. Then, the predicted label is determined by their goodness. The chosen label is the one with the highest overall goodness.

\begin{figure*}[t!]
     \centering\small
     \begin{subfigure}[b]{0.28\textwidth}
         \centering         \includegraphics[width=\textwidth]{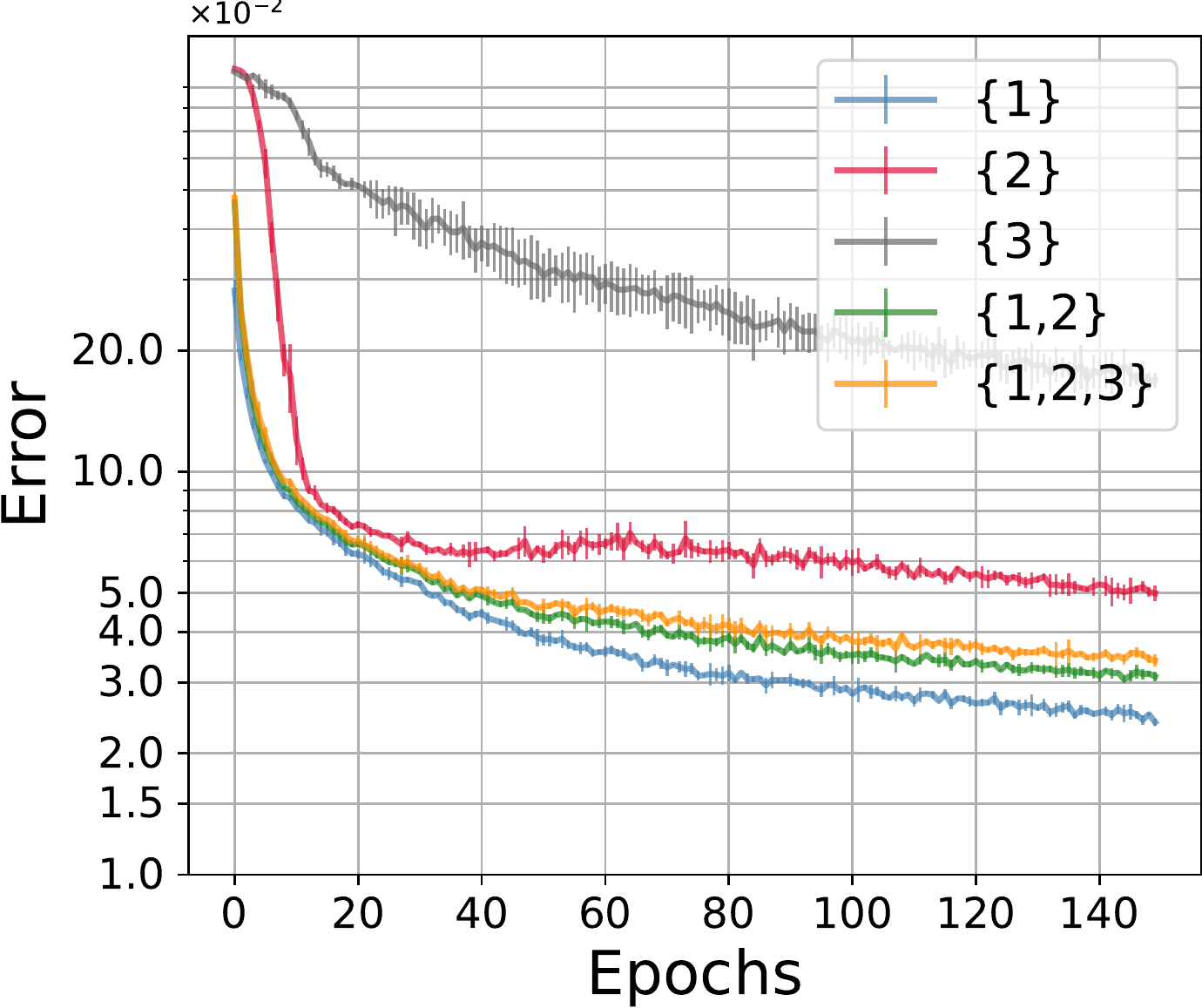}
         \caption{MNIST (FF)}
         \label{fig:layer_loss_coop_mnist}
     \end{subfigure}
     \begin{subfigure}[b]{0.26\textwidth}
         \centering
         \includegraphics[width=\textwidth]{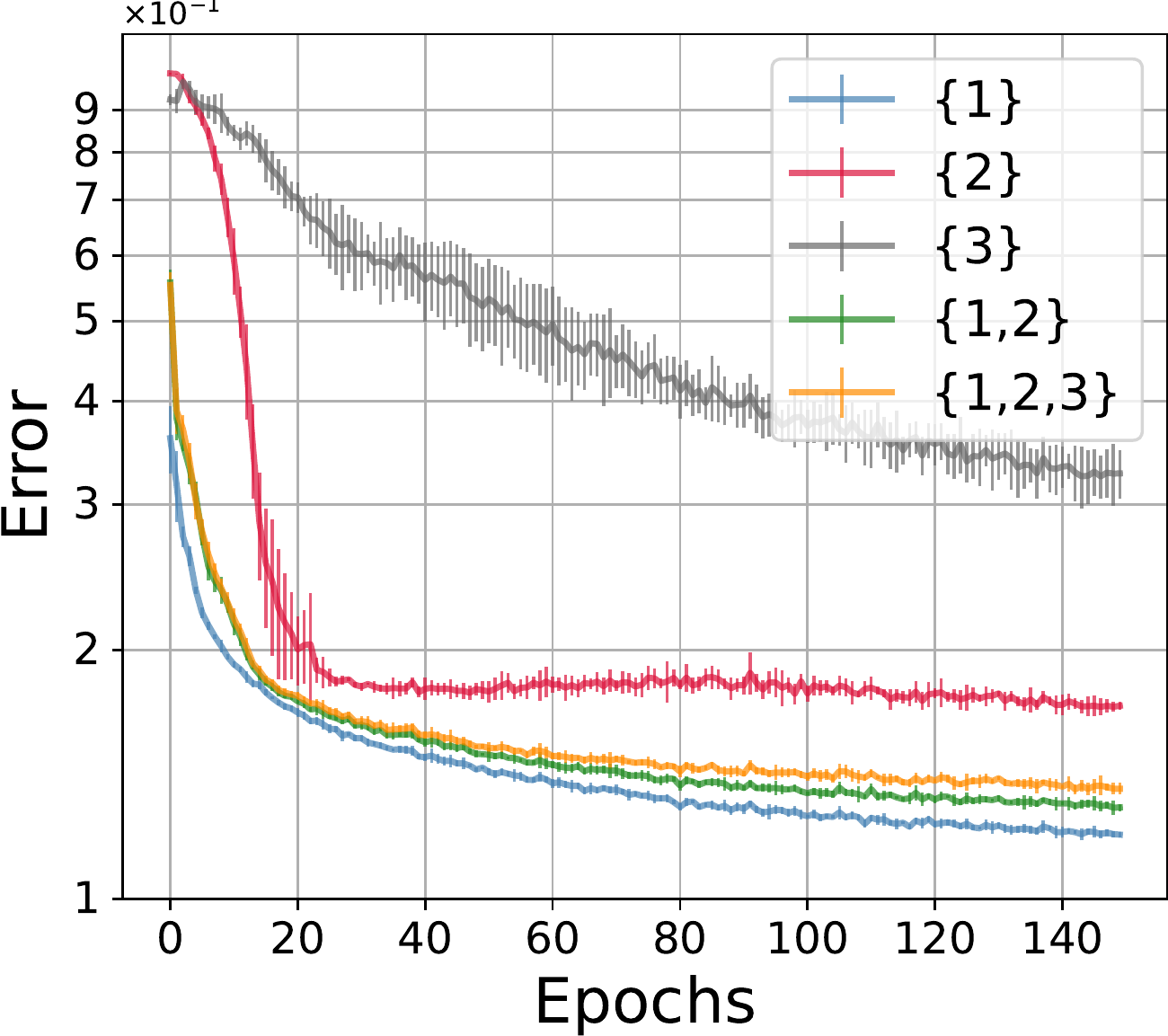}
         \caption{Fashion (FF)}
         \label{fig:layer_loss_coop_fashion}
     \end{subfigure}
     \begin{subfigure}[b]{0.26\textwidth}
         \centering
         \includegraphics[width=\textwidth]{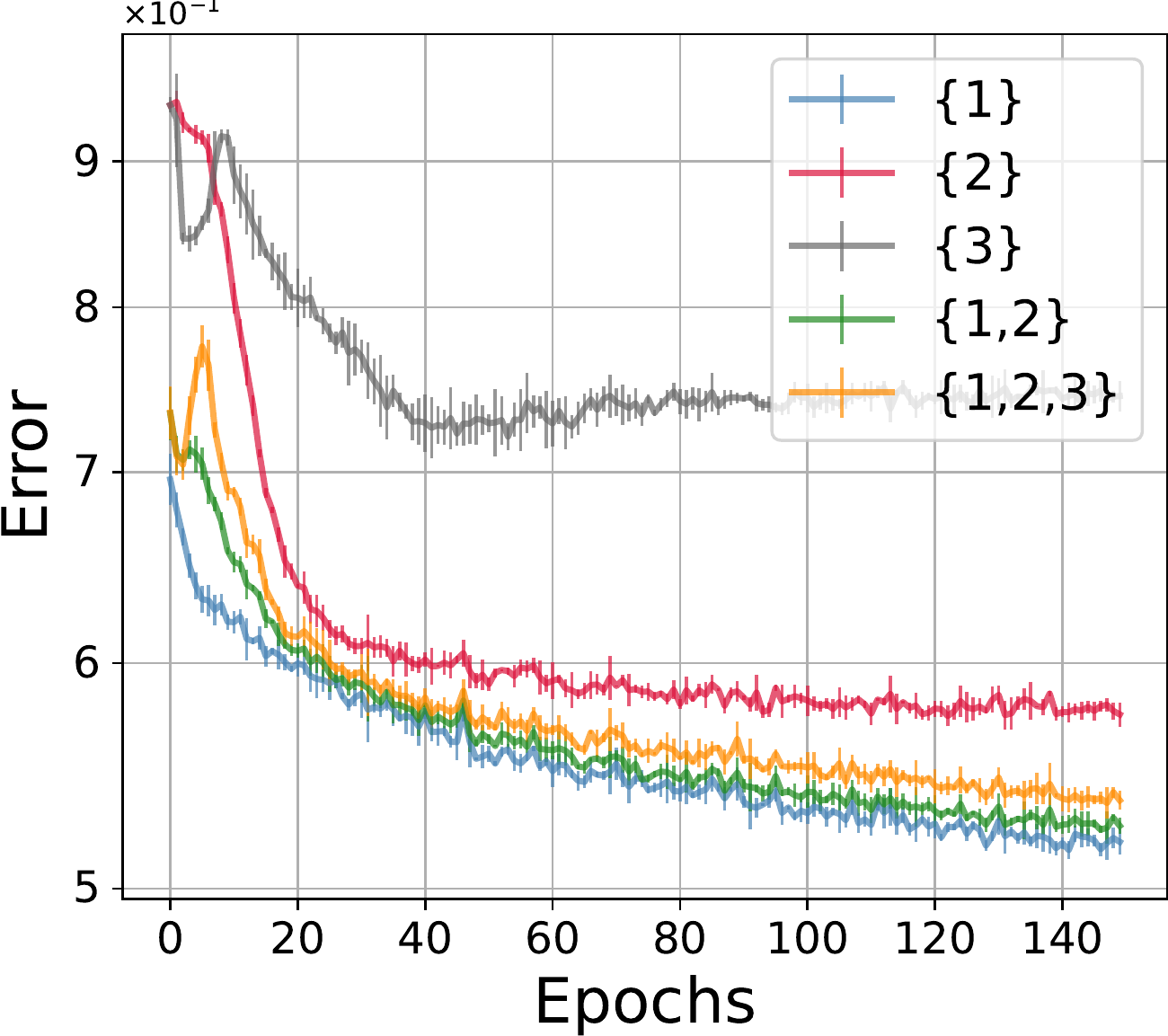}
         \caption{CIFAR-10 (FF)}
         \label{fig:layer_loss_coop_cifar}
     \end{subfigure}
     \vfill
     \begin{subfigure}[b]{0.28\textwidth}
         \centering
         \includegraphics[width=\textwidth]{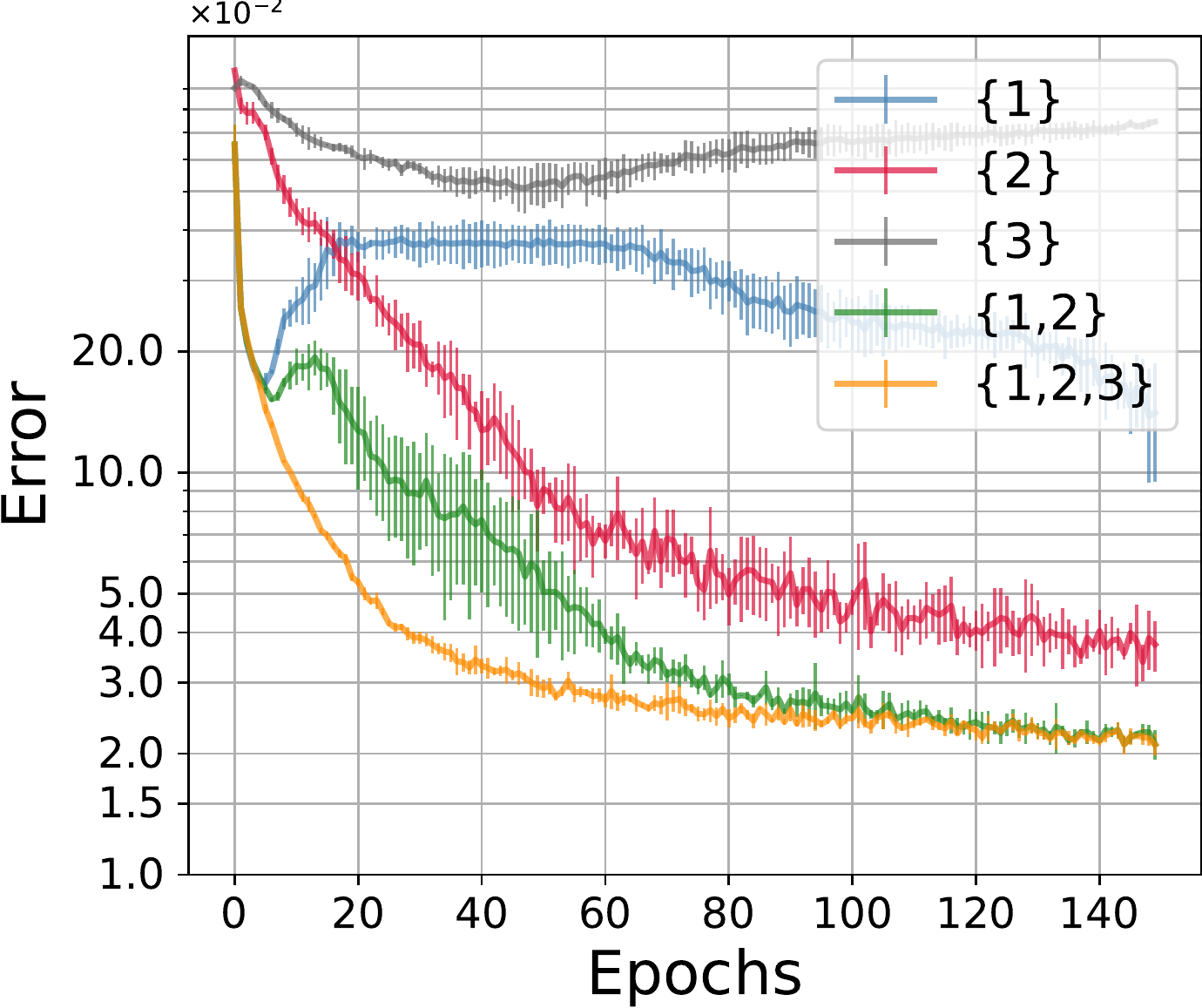}
         \caption{MNIST (Collaborative)}
         \label{fig:unified_loss_mnist}
     \end{subfigure}
     \begin{subfigure}[b]{0.26\textwidth}
         \centering
         \includegraphics[width=\textwidth]{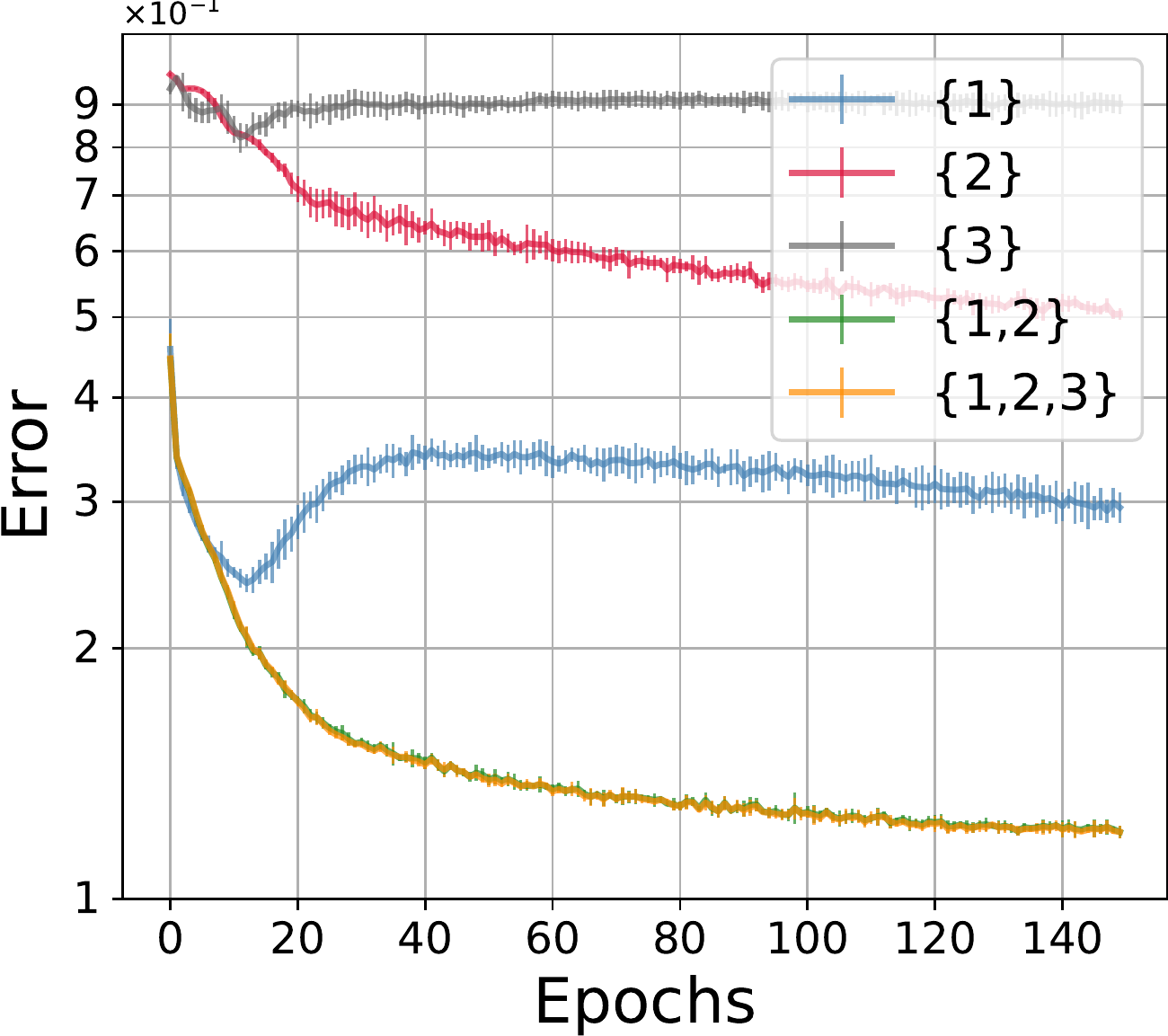}
         \caption{Fashion (Collaborative)}
         \label{fig:unified_loss_fashion}
     \end{subfigure}
     \begin{subfigure}[b]{0.26\textwidth}
         \centering
         \includegraphics[width=\textwidth]{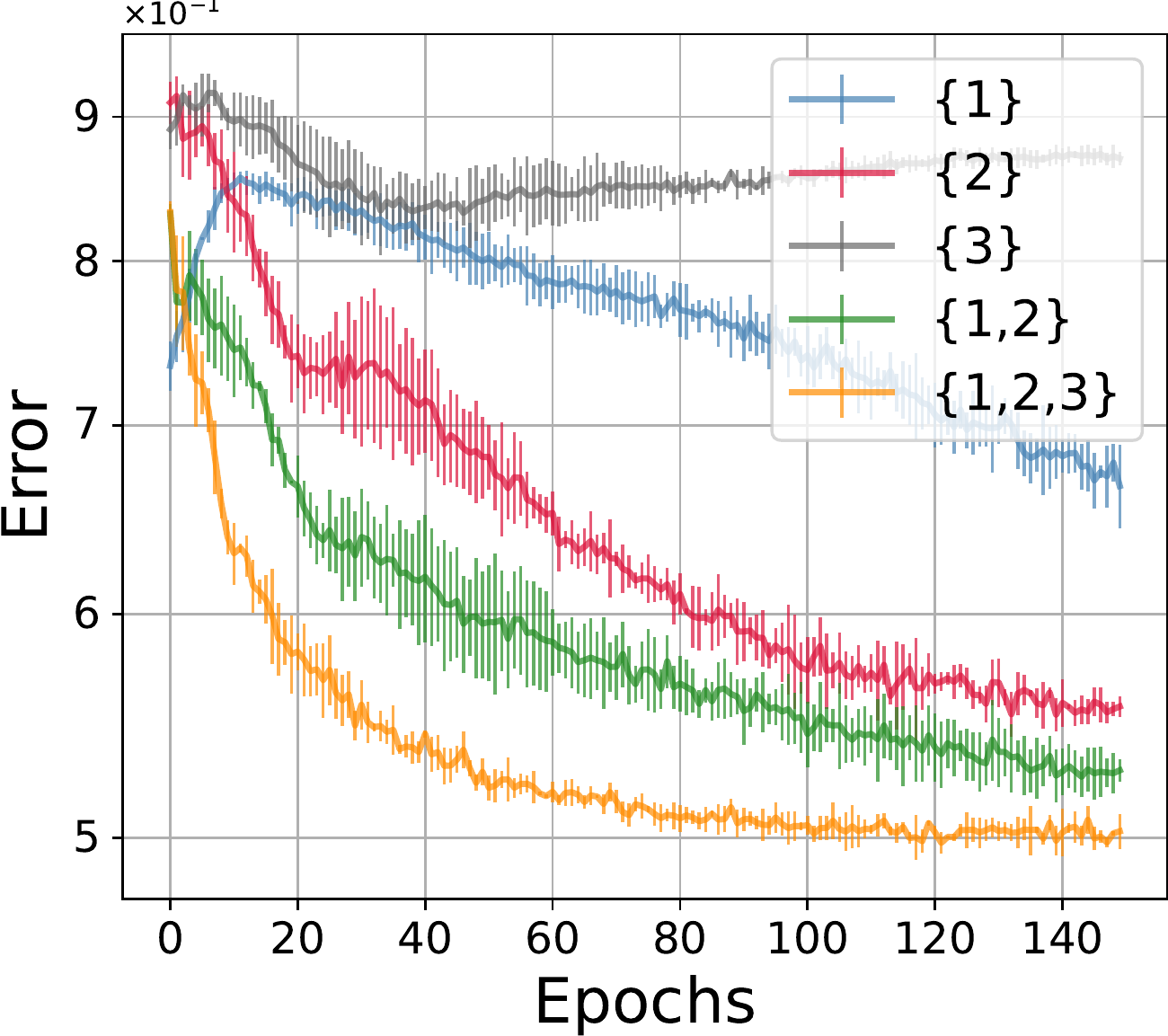}
         \caption{CIFAR-10 (Collaborative)}
         \label{fig:unified_loss_cifar}
     \end{subfigure}
     \caption{Layer Collaboration: Panels (a-c) show the results of individual and grouped layers. We observe that the original form of the forward-forward algorithm does not enable collaboration between layers. It is notable that the second and third layers do not contribute positively to performance, as their marginal contribution is negative. Panels (e-f) illustrate the effect of our proposed method on layer collaboration. Individually, each layer performs worse than layer loss. However, their marginal contribution is positive and their performance together outperforms the layer loss.}
     \label{fig:layer_coo}
\end{figure*}

\section{Layer collaboration}\label{sec:layer_colb}

Deep nets consist of multiple layers. Each layer processes the input data and passes it to the next layer, allowing the network to learn increasingly complex representations of the data. Thus, layer collaboration is crucial in deep nets because it allows the network to learn and extract features from the data in a hierarchical and efficient manner. For example, lower layers of the network might learn to identify `simple' features such as edges and textures, while later layers might learn complex features such as object shapes and patterns. This hierarchical structure allows the network to build a robust representation of the data and make accurate predictions. 

The forward-forward algorithm permits communication between layers only through the forward pass because each layer only takes into account the output of its predecessor. In this work, we find that the forward collaboration solely is not sufficient for developing a complex representation of the data. One can observe from Equation~\ref{eq:ff_step} that the forward-forward process detailed above does not enable the flow of information to earlier layers (i.e., layers closer to the data) while training (see Figure~\ref{fig:motivation}). For example, the first layer is not aware of the existence of later layers and hence can not consider information from them in its optimization process. This is contrary to the backpropagation algorithm. From Equation~\ref{eq:backprop} we observe that information can flow from a layer to its predecessors through the gradients of the optimization step.

\begin{figure*}[t!]
\begin{minipage}[t!]{0.48\textwidth}
\begin{algorithm}[H]
    \centering\small
    \caption{Forward-Forward}\label{algorithm:layer_loss}
    \begin{algorithmic}[1]
        \State \text{\textbf{Input:} $\theta, f, S = \{(x_1, y_1),..,(x_m, y_m)\}$}
        \For{$i\gets 1, k$}
        \Repeat
        \State \text{Sample $(x, y)$ from $S$}
        \State \text{\color{white} $\gamma \gets \sum_{t\ne i} \norm{\hat{f}_{1:t}(x,y)}^2$}
        \State $p \gets \left(1 + e^{-(\norm{f_{1:i}(x,y)}^2 - \theta)} \right)^{-1}$
        \State $w_i \gets w_i - \frac{\partial p}{\partial w_i}$ 
        \Until {Convergence}
        \EndFor
    \end{algorithmic}
\end{algorithm}
\end{minipage}
\hfill
\begin{minipage}[t!]{0.48\textwidth}
\begin{algorithm}[H]
    \centering\small
    \caption{Collaborative Forward-Forward}\label{algorithm:unified_loss}
    \renewcommand{\algorithmicrepeat}{\textcolor{myred}{\textbf{repeat}}}
    \renewcommand{\algorithmicfor}{\textcolor{myred}{\textbf{for}}}
    \renewcommand{\algorithmicend}{\textcolor{myred}{\textbf{end}}}
    \renewcommand{\algorithmicdo}{\textcolor{myred}{\textbf{do}}}
    \renewcommand{\algorithmicuntil}{\textcolor{myred}{\textbf{until}}}
    \begin{algorithmic}[1]
        \State \text{\textbf{Input:} $\theta, f, S = \{(x_1, y_1),..,(x_m, y_m)\}$}
        \Repeat
        \For{{\color{myred}$i\gets 1, k$}}
        \State \text{Sample $(x, y)$ from $S$}
        \State \text{\color{myred} $\gamma \gets \sum_{t\ne i} \norm{\hat{f}_{1:t}(x,y)}^2$}
        \State $p \gets \left(1 + e^{-(\norm{f_{1:i}(x,y)}^2 + {\color{myred} \gamma} - \theta)} \right)^{-1}$
        \State $w_i \gets w_i - \frac{\partial p}{\partial w_i}$ 
        \EndFor
        \Until {{\color{myred} Convergence}}
    \end{algorithmic}
\end{algorithm}
\end{minipage}
\captionof*{algorithm}{Algorithm Comparison: Difference between the classical and the proposed collaborative forward-forward. The collaborative approach adds a constant $\gamma$ to the probability computation. $\gamma$ modifies the value of $\theta$ by taking into account the progress of the training, while considering performance and layer collaboration. Note that during the optimization of a given layer, both versions compute gradients solely with respect to its weights. For readability, we show only positive optimization, the same holds for the negative step.}
\end{figure*}

In the following, we examine the collaboration between layers when training models using the forward-forward algorithm. We then propose a modification to the forward-forward algorithm that enables better collaboration between layers.

\subsection{Analysis} \label{sec:analysis}

The forward-forward inference process takes into account the sum of the goodnesses of the model's layers. This process can be viewed as a vote among the layers regarding the label $y$ that is most likely to fit the data sample $x$. Such an independent voting process allows us to examine the performance of each layer individually by considering their goodness. Similarly, one can evaluate the performance of subgroups of layers.

Figures~\ref{fig:layer_coo} (a-c) present for three datasets (MNIST, Fashion-MNIST, and CIFAR-10) the error of each layer individually and when they are combined to operate together. Those figures empirically verify that the forward-forward algorithm is unable to learn an intricate and collective representation. To see this, note that individually, each layer consistently performs better than its successor. In addition, together, the marginal contribution of each layer to the performance is negative (i.e., higher error) and the first layer alone achieves the lowest error. This is an undesired property for deep neural networks. This  implies that, in its current form, the forward-forward algorithm performs better using shallow networks, specifically, one layer networks.

Next, we propose a modification to the forward-forward algorithm that allows layers to collaborate regardless of their position in the network.

\subsection{Collaborative Forward-Forward}

The observations in Section~\ref{sec:analysis} encourage to seek a method that enables layers to better collaborate during training. As discussed before, breaking down the forward-forward training procedure reveals that early layers are not aware of their predecessors. In the following, we propose a simple, yet highly effective modification to the forward-forward algorithm. Concretely, we propose to consider the global goodness of the network for the optimization of each layer.

The threshold $\theta$ in the forward-forward algorithm governs the magnitude of the learning. To account for the global goodness of the network with respect to a given sample $(x,y)$, we propose to add the goodness of all layers to a given layer optimization step. Formally, we propose that the modified probability for positive samples reads as follows,
\[
    p_{i}(\text{positive}) \triangleq \left(1 + e^{-(\norm{f_{1:i}(x,y)}^2 + \gamma - \theta)} \right)^{-1},
\]
where, the \emph{constant} $\gamma$ is a sum of goodness values from different layers. For example,
\[
    \gamma = \sum_{t\ne i} \norm{\hat{f}_{1:t}(x,y)}^2 \quad\text{or}\quad \gamma_{<t} = \sum_{t< i} \norm{\hat{f}_{1:t}(x,y)}^2.
\]
Importantly, $\hat{f}$ is a copy of $f$ with fixed parameters (i.e., we do not take gradients with respect to $\hat{f}$). The proposed modification results in the following modified update step, with respect to $w_i$:
\[
     \Delta w_i &=& \frac{\partial \log\left(\left(1 + e^{-(\norm{f_{1:i}(x,y)}^2 + \gamma - \theta)} \right)^{-1}\right)}{\partial w_i}.\label{eq:unified}
\]
Such an update step allows information flow between layers. Intuitively, this ensures that the \textit{i}-th layer is aware of its later layers and its parameters are trained accordingly. It is important to note that our method maintains the original forward-forward assumptions and requirements.

\begin{wrapfigure}{r}{0.5\textwidth}
    \vspace{-10pt}
    \centering
    \includegraphics[width=0.5\textwidth]{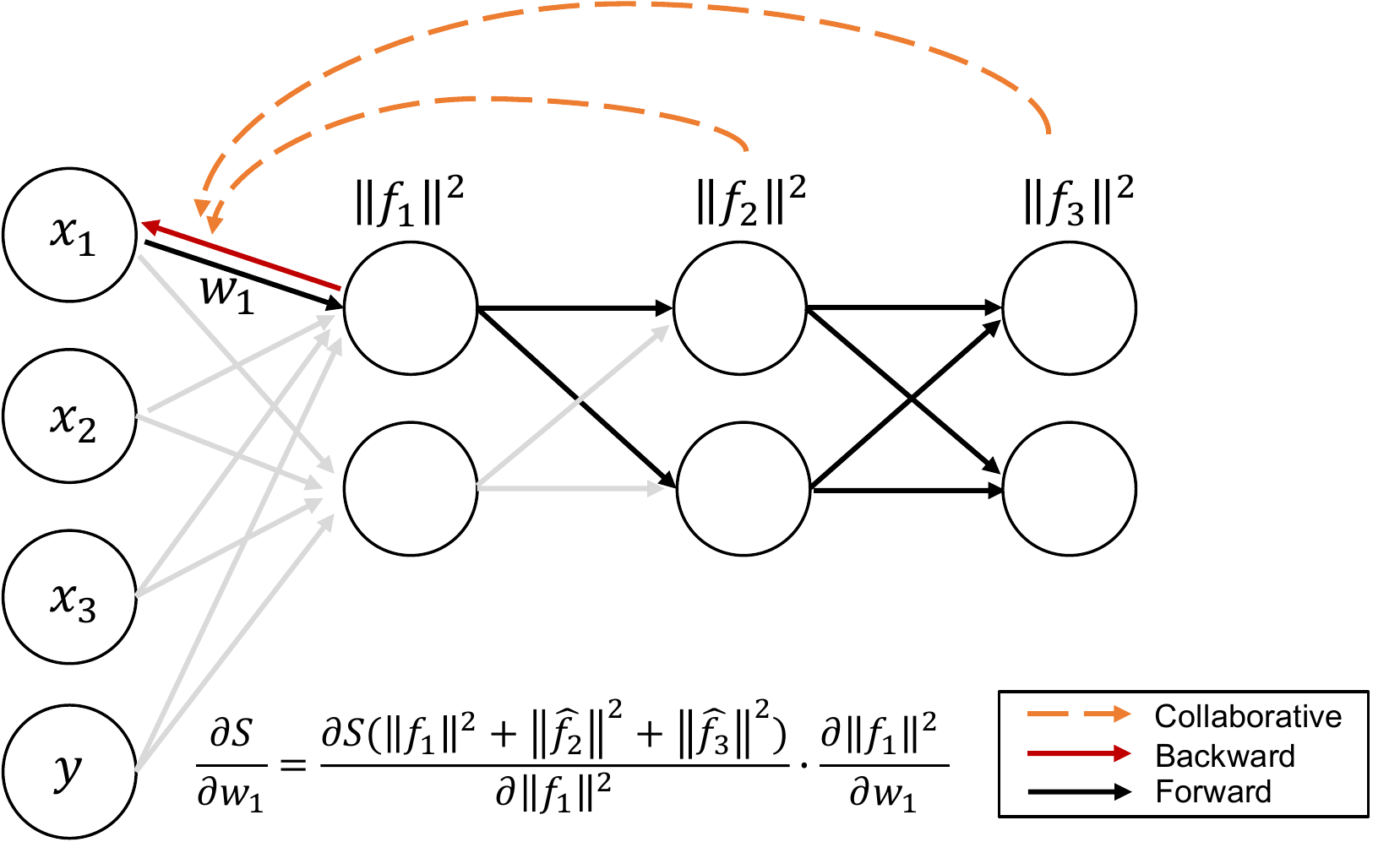}
    \caption{Illustration of our method: We incorporate and encourage a collaborative aspect among layers. By taking into account the goodness from all layers, the model is able to incorporate the information from future layers into its learning process.}
    \label{fig:ours_unified}
    \vspace{-10pt}
\end{wrapfigure}

We illustrate our method in Figure~\ref{fig:ours_unified}. The dashed lines represent $\hat{f}$, which governs the magnitude of the gradient. Algorithms~\ref{algorithm:layer_loss} and~\ref{algorithm:unified_loss} illustrate the difference between the layer and the unified loss. The layer loss requires to manually set the hyperparameter $\theta$, which turns out to be  sensitive and its value varies between tasks. Moreover, it is not clear how $\theta$ should be modified during training. Our method modifies the value of $\theta$ based on the progress of the training, while considering performance and collaboration between layers. 

We also suggest to modify the training procedure. The classical forward-forward algorithm trains each layer until convergence. Differently, we suggest to alternate the updates of all the layers' parameters. The combination of the unified loss and the modified training procedure yields a collaborative forward-forward method. We later show that this simple modification of the forward-forward algorithm is highly effective and improves results.

\section{Entropic view of layer collaboration}\label{sec:entropy}

The functional entropy can measure the amount of information that is shared across layers in the forward-forward algorithm, and consequently allows us to quantify and compare layer collaboration. In the following section we formulate the connection between the forward-forward algorithm and functional entropy. Using this connection, we show that during training, the functional entropy of the network implicitly improves. In Section~\ref{sec:experiments}, we propose to directly use the functional entropy during training and show that it is beneficial to the performance.

The functional entropy is defined over a non-negative random variable, which we denote by $h(z)$ that is defined over the space ${\cal Z}$ with a probability measure $\mu(z)$. Here and throughout, we use $z$ to refer to a stochastic variable, which we will integrate over. The functional entropy of the non-negative function $h(z) \ge 0$ is 
\[
\ent_{\mu}(h) \triangleq \E_{z\sim\mu}\left[h(z) \log\left(\frac{h(z)}{\E_{z\sim\mu}[h(z)]}\right)\right].\label{eq:ent}
\]
The functional entropy is non-negative, namely, $\ent_{\mu}(h) \ge 0$ and equals  zero only if $h(z)$ is a constant. 

Our settings are defined over the input space $(x,y)$ where $x \in \mathbb{R}^d$ and $y \in {\cal Y}$. In the following, we denote by $z$ the data-label tuple $(x,y)$ combined with the layer index $i \in \{1,...,k\}$. We use ${\cal Z} = \mathbb{R}^d \times {\cal Y} \times \{1,...,k\}$ to refer to its domain. The functional entropy can be intuitively considered as a scaled KL-divergence between the distribution of the input of a layer and the output of a layer $h(z) \triangleq \norm{f_{1:i}(x,y)}^2$. Formally, consider the prior and posterior distributions, $p(z), q(z)$, i.e.,    
\[
p(z) \triangleq d\mu(z), \quad\text{and}\quad
q(z) \triangleq \frac{p(z) h(z)}{\mathbb{E}_{z \sim \mu} [h(z)]}.
\]
One can verify that $q(z)$ is a probability density function since $q(z) \ge 0$ and $\mathbb{E}_{z \sim \mu} [q(z)] = 1$. The functional entropy can now be described as the scaled KL-divergence between the prior and posterior distributions above: 
\[
\ent_{\mu}(h) &=& \mathbb{E}_{z \sim \mu} [h(z)] \cdot KL(q || p).
\]
Thus, intuitively, the functional entropy is the amount of learning that is done by a function $h(z)$, measured by the KL-divergence from the prior distribution of the input $p(z)$ to the posterior distribution of the output $q(z) \propto p(z)h(z)$. 

\begin{figure*}[t!]
     \centering
     \begin{subfigure}[b]{0.245\textwidth}
         \centering
         \includegraphics[width=\textwidth]{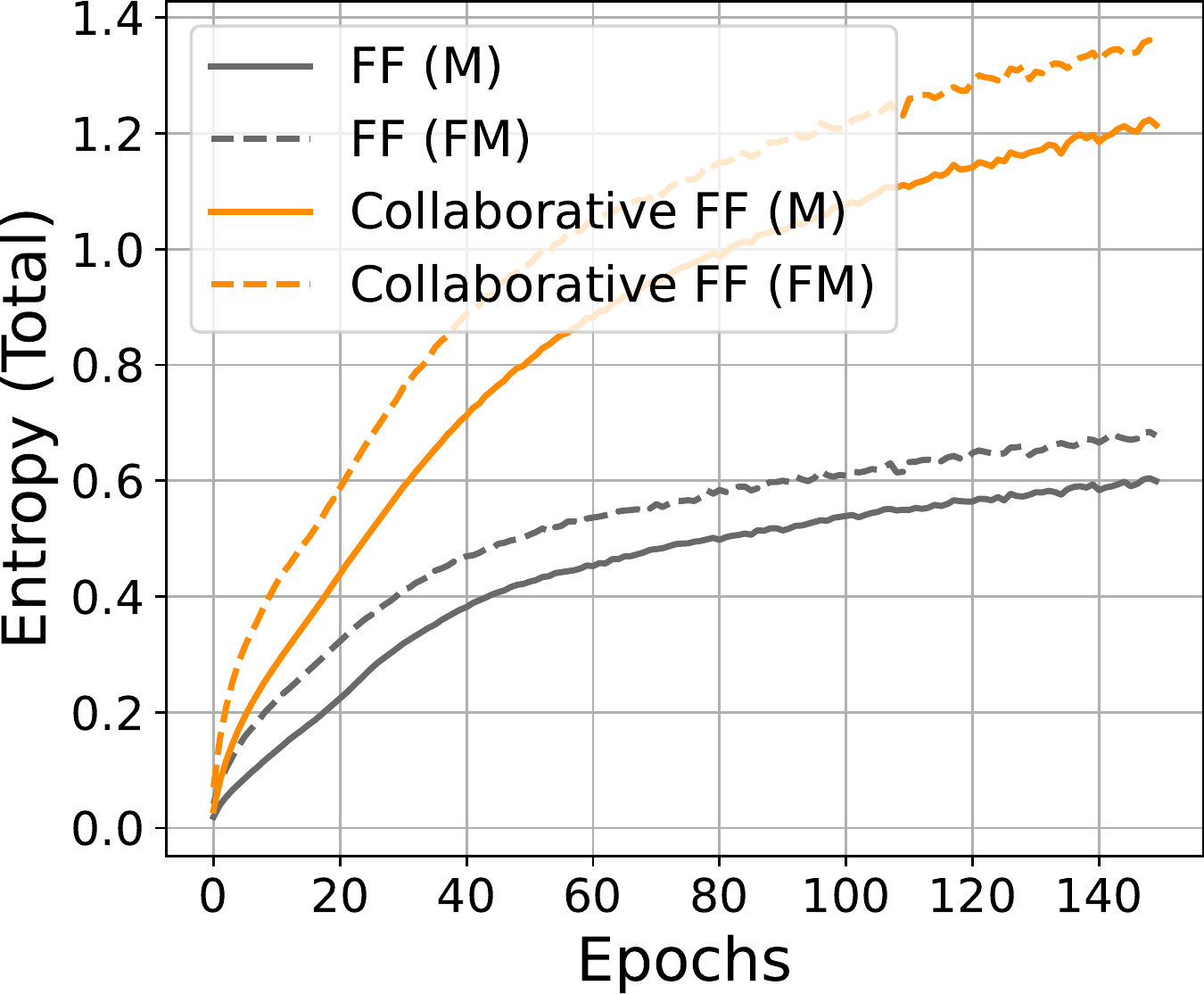}
         \caption{Overall}
         \label{fig:entropy_mnist}
     \end{subfigure}
     \begin{subfigure}[b]{0.245\textwidth}
         \centering
         \includegraphics[width=\textwidth]{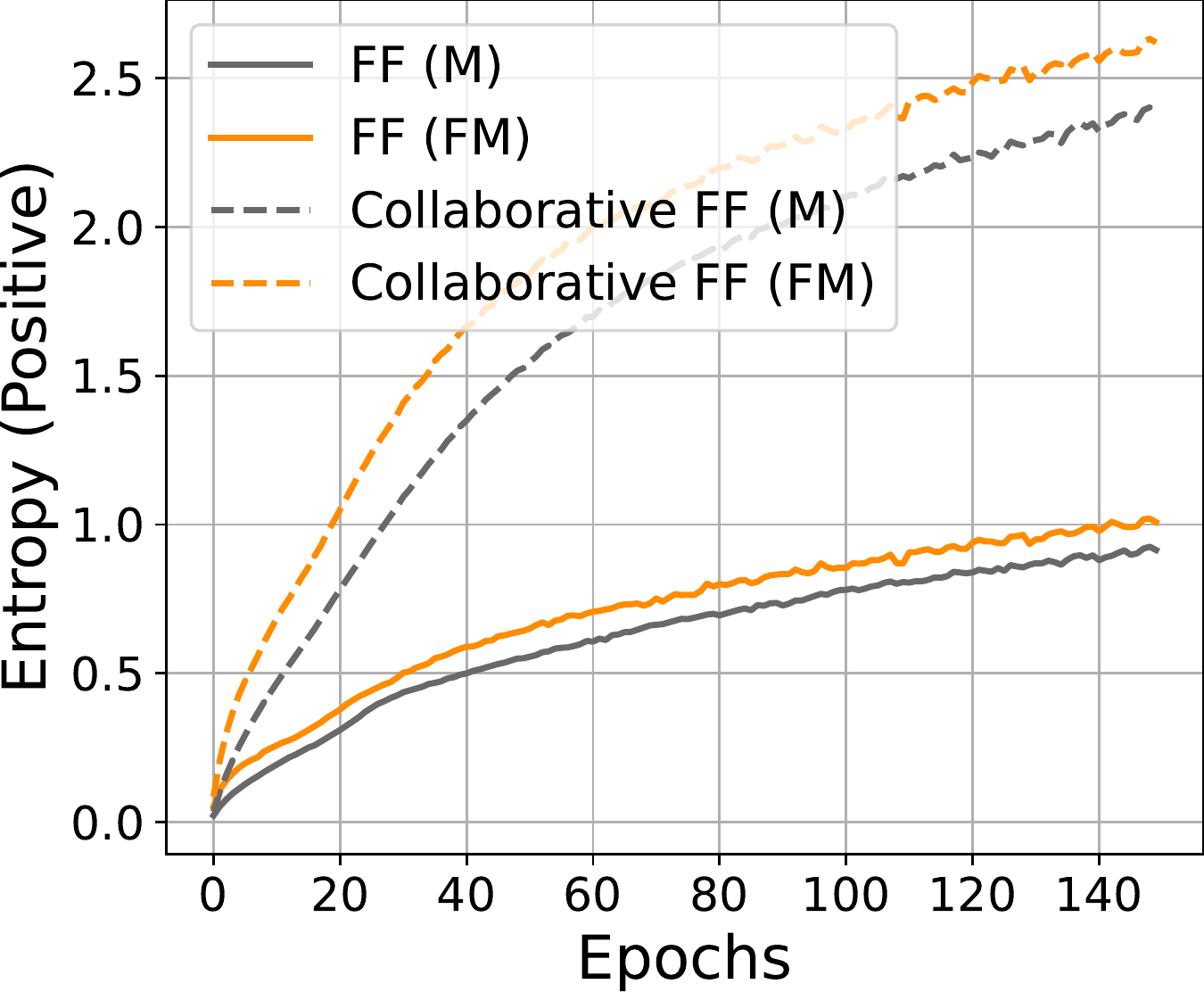}
         \caption{Positive Samples}
         \label{fig:entropy_positive}
     \end{subfigure}
     \begin{subfigure}[b]{0.25\textwidth}
         \centering
         \includegraphics[width=\textwidth]{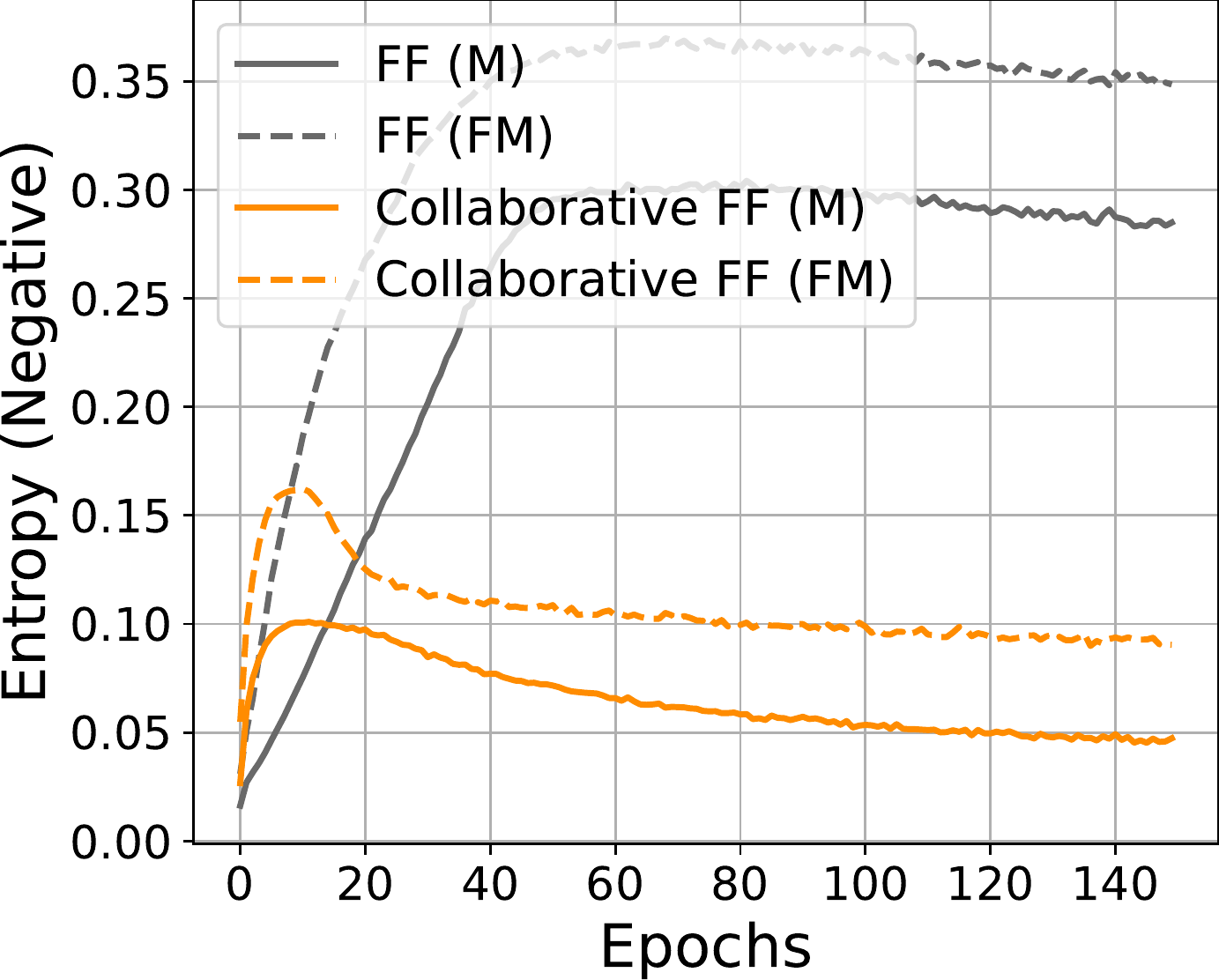}
         \caption{Negative Samples}
         \label{fig:entropy_negative}
     \end{subfigure}
        \begin{subfigure}[b]{0.235\textwidth}
         \centering
         \includegraphics[width=\textwidth]{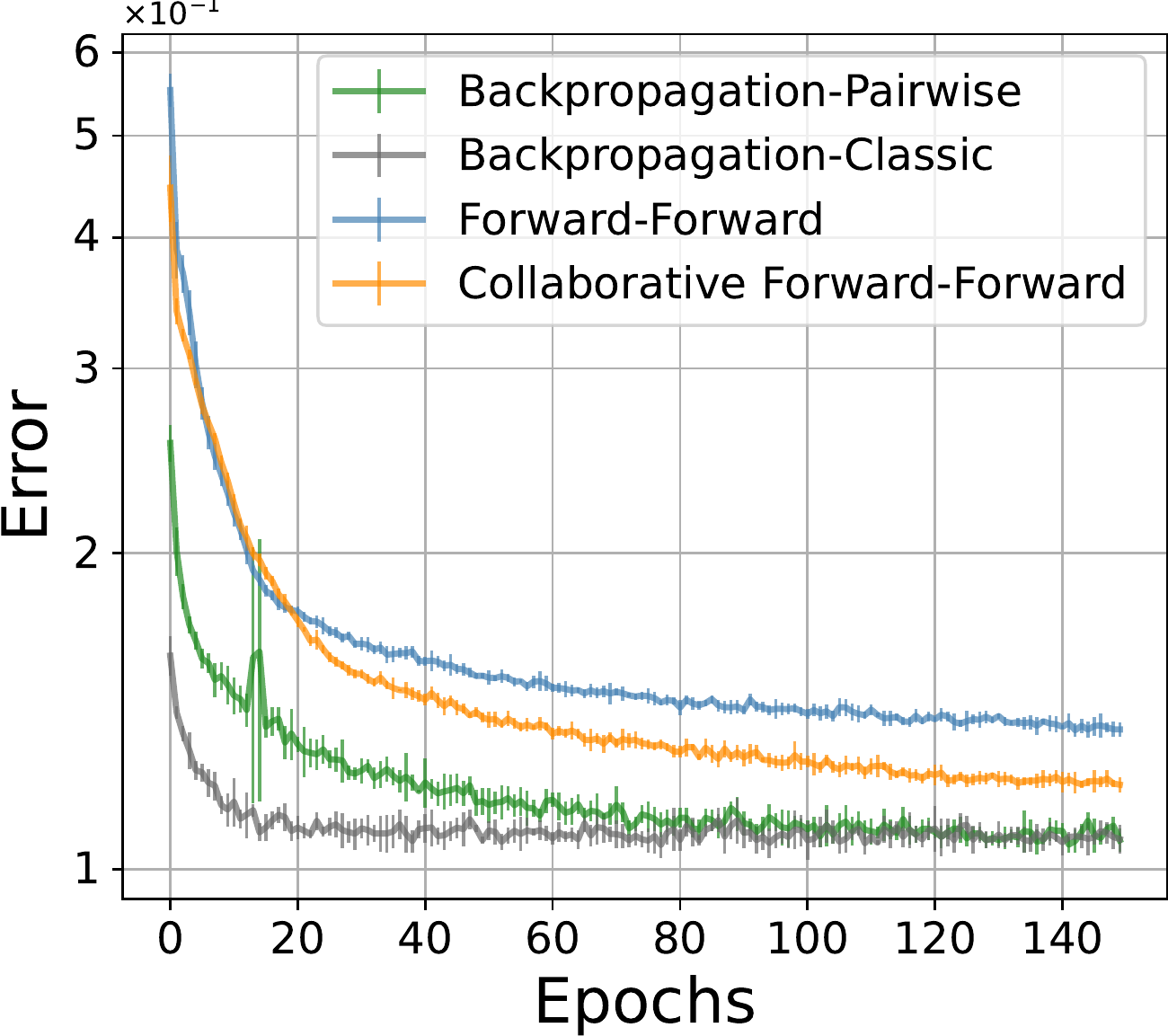}
         \caption{Performance}
         \label{fig:performance_comparison}
     \end{subfigure}
     \caption{Functional Entropy and Performance: We compare the functional entropy of classical and collaborative forward-forward using MNIST (M) and Fashion-MNIST (FM). In Panel (a) we measure entropy when considering both positive and negative samples during training. Panels (b) and (c) explore the entropy of the positive and negative samples individually (respectively). We observe that the forward-forward algorithm implicitly maximizes the functional entropy of the model. We also observe that the proposed unified loss attains a higher entropy than the layer loss. In Panel (d) we compare four models -- (1) `Forward-Forward' ; (2) `Collaborative Forward-Forward'; (3) `Backpropagation Pairwise' is trained using backpropagation to pairwise predict whether a concatenated sample ($z$) is positive or negative. This is in spirit similar to the forward-forward algorithm; (4) `Backpropagation Classic' is the classical backpropagation method.
     }
     \label{fig:func_ent}
\end{figure*}

In Figure \ref{fig:func_ent} we validate that our method improves the overall functional entropy of the learned network. We attribute its improvement to our learning algorithm, which encourages functional entropy to flow between the layers of the network. To formulate this notion, we decompose the functional entropy by  layers, and measure a  notion of conditional functional entropy with respect to the layers themselves. The probability measure $\mu(z)$ for $z = (x,y,i)$ accounts for the input $(x,y)$ of the $i$-th layer when computing its output $\norm{f_{1:i}(x,y)}^2$. The probability measure is $\mu(z) = \mu_1(i) \times \mu_2(x,y)$ which does not enforce statistical dependencies between the data generating distribution $\mu_1(x,y)$ and the layer distribution $\mu_2(i)$, for which we take the uniform distribution. In this setting, the functional entropy decomposes into two components\footnote{This decomposition can be verified by adding and subtracting $\E_{i} \left[\left( \E_{(x,y) | i} [h(x,y,i)] \right) \log \left( \E_{(x,y) | i} [h(x,y,i)] \right) \right]$ to $\ent_\mu[h]$.}: 
\begin{equation}
    \ent_{\mu}[h] = \ent_{\mu_1}\left[\E_{(x,y)}[h(x,y,\cdot)]\right] + \E_{i} \left[\ent_{\mu_2}[h(\cdot, \cdot, i)]\right].
\end{equation}
The first component is the functional entropy of the layers' distribution of the marginal function $\E_{(x,y)}[h(x,y,\cdot)]$. This entropy accounts for the information across layers. The seconds component is an average of entropies that are conditioned on the \textit{i}-th layer. 

\section{Results}\label{sec:experiments}

In this section, we study various modifications to the forward-forward algorithm. Specifically, we begin by studying the collaboration of layers when training using classical and collaborative forward-forward. We study the collaboration from both a performance and an entropy point of view. Then, we present the benefits of our method when using activations other than  ReLU. In addition, we find that the introduced functional entropy concept  is highly effective in replacing the sigmoid in forward-forward optimization.

\paragraph{Experimental setup} Similar to~\citet{ff}, we study all methods using  the MNIST~\citep{mnist} and CIFAR10~\citep{cifar10} datasets. We also study FashionMNIST~\citep{xiao2017fashion} as it is more challenging than MNIST. For all models, we use three layers with $500$ neurons in each layer. For the `Backpropagation Label Output' model, we project last hidden layer to the label space using an extra layer, while for discriminator-based models, the label is concatenated with the input data. All models are optimized using the Adam~\citep{kingma2014adam} optimizer. We apply a learning rate of $0.001$ and a batch size of $200$. All models are trained on a single GPU (NVIDIA 2080 Ti) for $150$ epochs.

\begin{table}[t!]\centering\small
\caption{Performance comparison of the classical forward-forward method, our proposed method (collaborative), backpropagation that predicts whether a concatenated sample is positive or negative, and classical backpropagation. The training column refers to the training procedure, `Layer' denotes training each layer to convergence, and `Network' refers to the procedure described in Algorithm~\ref{algorithm:unified_loss}. We compare those methods using MNIST, Fashion-MNIST, and CIFAR-10. Bold font highlights the best score. Our method outperforms the vanilla forward-forward method, but there is a gap to close to achieve backpropagation-like performance. The scores are averaged over five runs, all standard deviations are smaller than $0.2$.}
\begin{tabular}{llccc}
    \toprule
    Method  & Training & MNIST & Fashion-MNIST  & CIFAR-10  \\
    \midrule
    {\color{gray} Backpropagation Pairwise} &{\color{gray} Network}&{\color{gray} 2.0}& {\color{gray} 10.7} & {\color{gray} 45.4} \\
    {\color{gray} Backpropagation Label} &{\color{gray} Network}&{\color{gray} 1.5}& {\color{gray} 10.7} & {\color{gray} 45.9} \\
    \midrule
    Forward-Forward & Layer &3.3& 13.2 & 54.2 \\
    Forward-Forward& Network &3.4&13.6&53.7\\
    Collaborative Forward-Forward $\gamma_{< t}$ &Layer &2.2& \textbf{11.6} & 53.2 \\
    Collaborative Forward-Forward $\gamma_{< t}$ &Network &2.2& \textbf{11.6} & \textbf{51.6} \\
    Collaborative Forward-Forward $\gamma$ &Network &\textbf{2.1}& 12.0 & 52.7 \\
    \bottomrule
\end{tabular}
\label{table:performance}
\end{table}

\subsection{Layer collaboration}

This section covers collaboration between layers using various learning algorithms. We study information flow between layers when training a deep neural network with backpropagation, with forward-forward, and with our proposed modifications. In backpropagation-based models, early layers are updated according to their effect on later layers. Thus, they hierarchically learn features that are useful for the final output. In contrast, the forward-forward algorithm updates each layer separately without being aware of subsequent layers. This results in poor communication between layers and a limited information flow. Our results show that, in forward-forward-based models, the first layer on its own achieves the lowest prediction error and additional layers do not improve results further. This highlights the need for a modification that enables a better communication between layers. Our method allows layers to better take into account  global information. We study this form of layer collaboration of the aforementioned methods from a performance and a functional information point of view.

For performance reference, we provide backpropagation-based models results. The first model presented, employs classical backpropagation. It consists of several layers and its output equals the size of the label space. The second model is trained to replicate the forward-forward methodology. Concretely, the input is the sample $x$ concatenated with a label $y$. The model's task is to discriminate between positive and negative samples, it is done by pairwise ranking of concatenated samples~\citep{liu2009learning}. In this model, in each optimization step, the last layer's norm is maximized for positive samples and minimized for negative samples.

\paragraph{Performance.} Table~\ref{table:performance} summarizes the performance of the evaluated methods. Under the forward-forward setup, the proposed approach performs the best, being superior to the classical forward-forward under all evaluated setups. We also perform ablation study for different choices of $\gamma$ and training procedures. As expected, the backpropagation approach, in both classical and pairwise, reaches the best overall performance. Interestingly, optimizing the network using classical and pairwise objectives reaches comparable performance. 

\paragraph{Entropy.} Next, we analyze the total entropy of the network for both classical and collaborative forward-forward, considering MNIST and Fashion-MNIST. Results are depicted in Figure~\ref{fig:func_ent}. Results are reported for both positive and negative samples (Figure~\ref{fig:entropy_mnist}), positive samples (Figure~\ref{fig:entropy_positive}), and negative samples (Figure~\ref{fig:entropy_negative}). Results suggest the proposed method better maximizes the entropy across all settings. 

When considering the entropy of the negative samples solely, the entropy is minimized comparing to the classical forward-forward. This follows naturally from Equations~\ref{eq:ff_step} and~\ref{eq:unified}.

\subsection{Entropy-based optimization}

In Section~\ref{sec:entropy} we observe that the forward-forward algorithm implicitly maximizes the entropy of the network. This raises the question of whether directly optimizing the entropy will boost performance. We formulate the learning process as an entropy maximization process. For the classical forward-forward optimization, this results in maximizing the following objective for the \textit{i}-th layer:
\begin{equation}
    \ent_{\mu_2}[h(\cdot, \cdot, i)] = \E\left[h(x, y, i) \log\left(\frac{h(x, y, i)}{\E[h(x, y, i)]}\right)\right].    
\end{equation}
Note, $i$ is fixed, and the expectation is taken with respect to $(x, y)\sim\mu_2$. It is straightforward to modify the above equation to the unified loss. It can be done similar to Equation~\ref{eq:unified}, i.e., by adding the goodness of all other layers without considering their gradients.

This proposed optimization process, which does not include the sigmoid function, nor $\theta$ obtains on-par results to the classical and collaborative forward-forward. For the Fashion-MNIST dataset, considering layer optimization, it obtains an error of $13.7\%$ (vs. the original $13.6\%$) and for the collaborative version it obtains $12.9\%$ (vs. $12.0\%$).

\section{Related work}\label{sec:related}

Our work focuses on the  collaboration between layers in  forward-forward-based models. In this section, we review research directions that are related to the forward-forward method, layer collaboration in deep neural networks, and functional entropy.

\paragraph{Forward-Forward algorithm.} The forward-forward algorithm assigns a score to each $(x, y)$ pair using the goodness function. This process is similar to  energy-based models (EBMs) which assign an energy to each similar pair. EBMs are a class of generative machine learning models that can to model complex probability distributions~\citep{lecun2005loss, lecun2006tutorial}. They are based on the idea that the probability of a particular configuration of the network is determined by its energy. The energy is defined as a function of the network's parameters, and the probability of a configuration given by the Boltzmann distribution, which is proportional to the exponential of the negative energy. Energy-based models have been used in a variety of applications, such as image generation, natural language processing, and reinforcement learning~\citep{ngiam2011learning, liu2017learning, haarnoja2017reinforcement, sharma2021maximum}. One of the most popular types of energy-based models is the Boltzmann machine~\citep{salakhutdinov09a}. 

Another line of related work discusses the Restricted Boltzmann Machine (RBM)~\citep{fischer2012introduction}, which, similarly to the forward-forward algorithm, trains the network layer-by-layer. A notable difference: RBMs are unsupervised learning models, while the forward-forward method is a fully supervised.

\paragraph{Layer collaboration.} Various works have studied features learned using neural networks. Bottom-up approaches observe that networks start with simple features and build up to more complex ones. For example, in image recognition tasks, the network may first learn to recognize edges, then shapes, and finally objects~\citep{workman2015location, liu2021bottom}. In top-down approaches,  the network starts with high level features and then breaks them down into lower level ones. In these approaches, high level features are used to guide the learning of lower level features. For example, in natural language processing tasks, the network may use high level semantic features to guide the learning of lower level syntactic features~\citep{maulud2021state, adel2015syntactic}. Both bottom-up and top-down approaches have been used in combination in several works, e.g., by~\citep{mahdi2019deepfeat, anderson2018bottom}. These works have demonstrated that using high level features can help the network to learn more efficiently and improve the performance of the model. In general, the high-level features learned by the network are more abstract and semantically meaningful, while low-level features are more specific to the task at hand. This type of hierarchical representation learning is possible only when layers can collaborate with each other.

\paragraph{Functional entropy.} Functional entropy is gaining an increasing amount of attention from the machine learning community~\citep{jahnel2023trajectorial, gat2022latent, anceschi2021spatially, gat2022functional, Hypocoercivity}.~\citet{gat2022importance} used it to study the importance of the gradient norm for a PAC-Bayes generalization bound. Different from our work, the authors are able to study gradient flow in the network as it is differentiable.~\citet{gat2020removing} maximize entropies of data modalities as a regularization of biased datasets. This is different from our work since they operate over differentiable models and data modalities (i.e., not parameters). In this work, we find the functional entropy to be useful as a guiding concept for collaboration between layers in forward-forward-based models.

\section{Conclusions and limitations}

This work studies layer collaboration of models that are trained using the forward-forward algorithm. We find that, in its current form, the forward-forward algorithm is limited in its collaborative capabilities, meaning information can not propagate backward throughout the network. We study this phenomenon and propose a method to overcome this challenge. We further propose an entropic-based view of the forward-forward algorithm, and, through extensive experiments, show the benefits of following the proposed method.

Nevertheless, the forward-forward algorithm is far from being comparable to the standard backpropagation. Many questions remain open as to how it can be applied to complex applications, e.g., image classification, text analysis, generative learning, and modern neural architectures, e.g., convnets, transformers, etc. We hope the community will keep pursing such research directions and challenge popular methods.

\bibliographystyle{unsrtnat}
\bibliography{bib}

\end{document}